%% file: sn-article.tex
\documentclass[sn-mathphys-num]{sn-jnl}

\usepackage{graphicx}%
\usepackage{multirow}%
\usepackage{amsmath,amssymb,amsfonts}%
\usepackage{amsthm}%
\usepackage{bbm}     
\usepackage{amsmath}
\usepackage[title]{appendix}%
\usepackage{xcolor}%
\usepackage{textcomp}%
\usepackage{manyfoot}%
\usepackage{booktabs}%
\usepackage{algorithm}%
\usepackage{algorithmicx}%
\usepackage{algpseudocode}%
\usepackage{listings}%
\usepackage{subcaption}
\usepackage{adjustbox}
\usepackage{xcolor}
\usepackage[normalem]{ulem}
\usepackage{comment}
\usepackage{array}
\usepackage{tikz}
\usetikzlibrary{shapes.geometric, arrows}
\tikzstyle{materials} = [rectangle, rounded corners, minimum width=3em, minimum height=1.25em,text centered, draw=black, fill=red!20]
\tikzstyle{objects} = [rectangle, rounded corners, minimum width=3em, minimum height=1.25em,text centered, draw=black, fill=blue!20]
\tikzstyle{discipline} = [rectangle, rounded corners, minimum width=3em, minimum height=1.25em,text centered, draw=black, fill=orange!20]
\tikzstyle{subjects} = [rectangle, rounded corners, minimum width=3em, minimum height=1.25em,text centered, draw=black, fill=magenta!20]

\useunder{\uline}{\ul}{}
\tikzstyle{arrow} = [thick,->,>=stealth]
\newcommand{\etal}{\textit{et al}., }
\newcommand{\ie}{\textit{i}.\textit{e}., }

\newcommand{\minisection}[1]{\vspace{0.075in}\noindent {\bf #1}\ }

\begin{document}

\title{EUFCC-340K: A Faceted Hierarchical Dataset for Metadata Annotation in GLAM Collections}


\author[1]{\fnm{Francesc} \sur{Net}}\email{fnet@cvc.uab.cat}

\author[2]{\fnm{Marc} \sur{Folia}}\email{marc.folia@nubilum.es}

\author[2]{\fnm{Pep} \sur{Casals}}\email{pep.casals@nubilum.es}

\author[3]{\fnm{Andrew} \sur{D. Bagdanov}}\email{andrew.bagdanov@unifi.it}

\author[1]{\fnm{Lluis} \sur{Gómez}}\email{lgomez@cvc.uab.cat}

\affil[1]{\orgdiv{Computer Vision Center}, \orgname{Universitat Autònoma de Barcelona}, \orgaddress{\city{Barcelona}, \postcode{08290}, \state{Catalunya}, \country{Spain}}}

\affil[2]{\orgdiv{Nubilum}, \orgaddress{\street{Gran Via de les Corts Catalanes 575, 1r 1a}, \city{Barcelona}, \postcode{08011}, \state{Catalunya}, \country{Spain}}}

\affil[3]{\orgdiv{Media Integration and Communication Center}, \orgname{University of Florence}, \orgaddress{\city{Florence}, \postcode{50134}, \country{Italy}}}

\abstract{In this paper we address the challenges of automatic metadata annotation in the domain of Galleries, Libraries, Archives, and Museums (GLAMs) by introducing a novel dataset, EUFCC-340K, collected from the Europeana portal. Comprising over 340,000 images, the EUFCC-340K dataset is organized across multiple facets -- Materials, Object Types, Disciplines, and Subjects -- following a hierarchical structure based on the Art \& Architecture Thesaurus (AAT). We developed several baseline models, incorporating multiple heads on a ConvNeXT backbone for multi-label image tagging on these facets, and fine-tuning a CLIP model with our image-text pairs. Our experiments to evaluate model robustness and generalization capabilities in two different test scenarios demonstrate the dataset's utility in improving multi-label classification tools that have the potential to alleviate cataloging tasks in the cultural heritage sector. The EUFCC-340K dataset is publicly available at \url{https://github.com/cesc47/EUFCC-340K}.}


\maketitle

\section{Introduction}\label{sec:introduction}
GLAMs (Galleries, Libraries, Archives, and Museums) are institutions that compile and maintain collections of diverse cultural heritage assets in the public interest. Their main functions are to preserve assets with cultural and historical value and make them available to researchers and the broader public alike. However, the traditional framework for asset cataloging within GLAMs relies heavily on manual metadata annotation, a complex and labor-intensive process undertaken by expert professionals adhering to stringent standards. This method, while effective, presents challenges to scalability and efficiency.

The motivation behind our work lies in the recognition of these challenges and the need for innovative solutions. We aim to take develop advanced tools that streamline the cataloging workflow and make a step forward in the use of artificial intelligence to help GLAM cataloging experts in the inventory process for cultural heritage collections. 

\begin{figure}
  \centering
  \begin{tabular}{ccc}
  \toprule
     \includegraphics[height=0.29\linewidth]{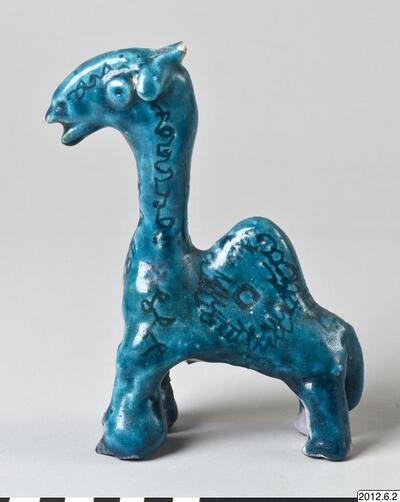}  & \includegraphics[height=0.29\linewidth]{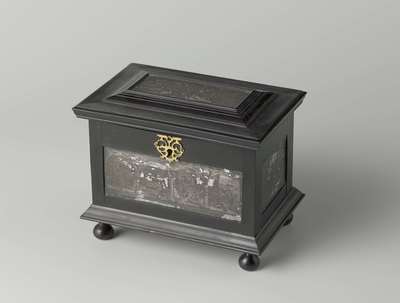}& \includegraphics[height=0.29\linewidth]{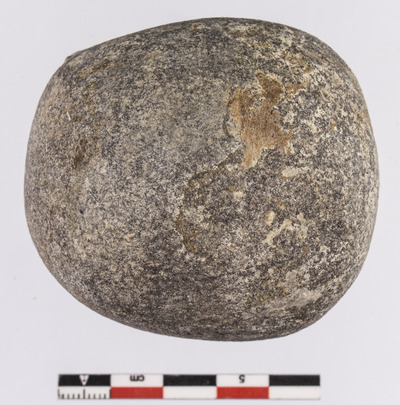} \\
\midrule
     Pottery  & Furniture & Tool \\
\midrule
\begin{tikzpicture}[node distance=1em]
\node (ObjectType) [objects] {\tiny{ObjectType}};
\node (Ceramics) [objects, below of=ObjectType, yshift=-0.85em] {\tiny{Pottery}};
\node (Figure) [objects, below of=Ceramics, yshift=-0.85em] {\tiny{Figure}};
\node (Material) [materials, right of= ObjectType, xshift=3em] {\tiny{Material}};
\node (Clay) [materials, below of= Material, yshift=-0.85em] {\tiny{Clay}};
\node (Ceramics2) [materials, below of= Clay, yshift=-0.85em] {\tiny{Ceramics}};
\draw [arrow] (ObjectType) -- (Ceramics);
\draw [arrow] (Ceramics) -- (Figure);
\draw [arrow] (Material) -- (Clay);
\draw [arrow] (Clay) -- (Ceramics2);
\node (nul) [below of=Ceramics2, yshift=-1.15em] {};  
\end{tikzpicture}
&
\begin{tikzpicture}[node distance=1em]
\node (ObjectType) [objects] {\tiny{ObjectType}};
\node (Ceramics) [objects, below of=ObjectType, yshift=-0.85em] {\tiny{Furniture}};

\node (Material) [materials, right of= ObjectType, xshift=4em] {\tiny{Material}};
\node (Clay) [materials, below of= Material, yshift=-0.85em] {\tiny{Wood}};
\node (Ceramics2) [materials, below of= Clay, yshift=-0.85em] {\tiny{Hard wood}};
\node (Ceramics3) [materials, below of= Ceramics2, yshift=-0.85em] {\tiny{Ebony}};
\draw [arrow] (ObjectType) -- (Ceramics);

\draw [arrow] (Material) -- (Clay);
\draw [arrow] (Clay) -- (Ceramics2);
\draw [arrow] (Ceramics2) -- (Ceramics3);
\end{tikzpicture}
&
\begin{tikzpicture}[node distance=1em]
\node (ObjectType) [objects] {\tiny{ObjectType}};
\node (Ceramics) [objects, below of=ObjectType, yshift=-0.85em] {\tiny{Tools \& equipment}};
\node (Figure) [objects, below of=Ceramics, yshift=-0.85em] {\tiny{Tool}};
\node (Material) [materials, right of= ObjectType, xshift=4em] {\tiny{Material}};
\node (Clay) [materials, below of= Material, yshift=-0.85em] {\tiny{Rock}};
\node (Ceramics2) [materials, below of= Clay, yshift=-0.85em] {\tiny{Igneous rock}};
\node (Ceramics3) [materials, below of= Ceramics2, yshift=-0.85em] {\tiny{Basalt}};
\draw [arrow] (ObjectType) -- (Ceramics);
\draw [arrow] (Ceramics) -- (Figure);
\draw [arrow] (Material) -- (Clay);
\draw [arrow] (Clay) -- (Ceramics2);
\draw [arrow] (Ceramics2) -- (Ceramics3);
\end{tikzpicture}
\\
\bottomrule
  \end{tabular}
     
  \caption{Sample records from the EUFCC-340K dataset. Each image in the dataset is annotated across different facets of the Getty ``Art \& Architecture Thesaurus'' (Some nodes are omitted for visualization purposes in this figure). }
  \label{image:europeana}
\end{figure}
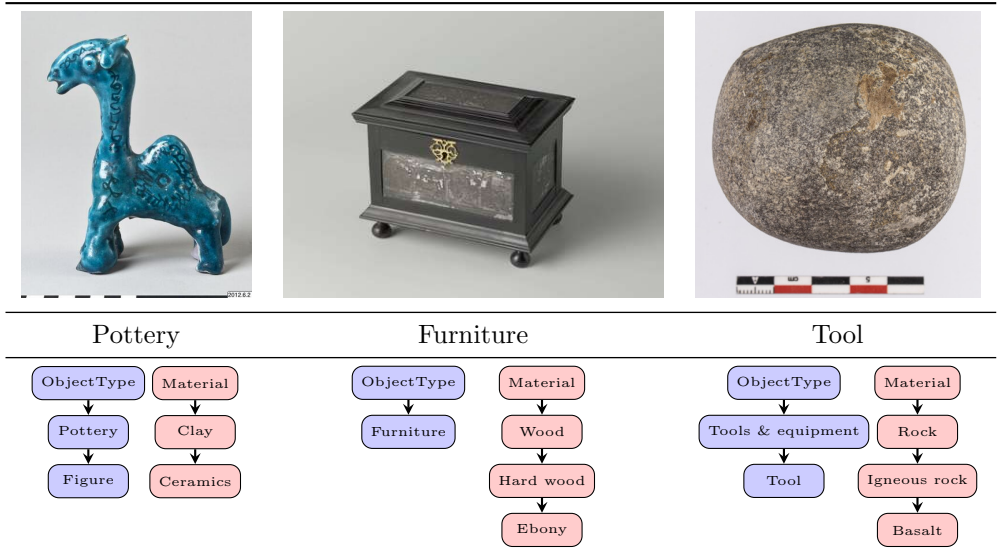

To this end we take an image tagging perspective. Image tagging consists of automatically assigning a list of tags (or category labels) to a given cultural asset's image, referring to words that describe its visual content and/or cultural context. This is a multi-label classification task in which an image usually belongs to more than one class (i.e. class labels are not mutually exclusive).

However, using state-of-the-art neural networks for automatic metadata annotation in the GLAM domain comes with its own set of challenges. First, the number of categories we need to consider can be very large -- for example, the Getty Research Institute's ``Art \& Architecture Thesaurus''\footnote{\url{https://www.getty.edu/research/tools/vocabularies/aat/}} has more than $165,000$ terms -- and the number of training examples for most of them can be very small (long tail)~\cite{van2017devil,wu2023automated}. On the other hand, the problem of incomplete annotations or missing labels is caused by the fact that two human annotators may consider different subsets of labels relevant for the same image even though both annotations may be equally valid~\cite{veit2018separating}. 

To address these challenges, we have created a novel dataset specifically tailored for image tagging in the GLAM domain. Our dataset, named EUFCC-340K (Europeana Facets Creative Commons), was collected from the Europeana portal\footnote{\url{http://www.europeana.eu}}, which aggregates a diverse range of GLAM European institutions, encompassing galleries, libraries, archives, and museums. EUFCC-340K comprises a total of $383,695$ images of a rich variety of cultural heritage assets, including artworks, sculptures, numismatics artifacts, textiles and costumes, and more, reflecting the diverse nature of collections within GLAM institutions. As illustrated in Figure~\ref{image:europeana} each image in the dataset is accompanied by manual annotations provided by domain experts, capturing a wide spectrum of descriptive tags.

In addition to the rich visual content of the images, the annotations provided for each image follow a structured framework consisting of several key facets. These facets include Materials, Object Types, Disciplines, and Subjects, each representing distinct aspects of the cultural heritage assets depicted in the images. Importantly, the terms used for annotation within each facet adhere to a hierarchical organization provided by the ``Art \& Architecture Thesaurus'' (AAT). In this way, we leverage the domain knowledge encapsulated within the thesaurus, thereby enhancing the interpretability and utility of our dataset. 

The hierarchical structure of the dataset annotations provides inherent advantages in mitigating challenges associated with sparse training data for certain categories in the long-tail, particularly those situated at the leaf nodes of the hierarchy. Thanks to the hierarchical organization, the dataset facilitates more robust predictions for broader categories at higher levels of the hierarchy, even when training data for specific subcategories may be limited. This characteristic not only enhances the applicability of the dataset in real-world use cases but also contributes to the effectiveness of machine learning models in handling the long tail distribution of metadata annotations within the GLAM domain. 

To evaluate the effectiveness of our dataset and to benchmark the performance of multi-label classification models in the GLAM domain, we also have developed a set of baselines tailored to address the multifaceted nature of this task. Our approach extends a state-of-the-art image classification architecture by augmenting them with multiple heads, each dedicated to classifying images based on a distinct facet within our annotation framework. Specifically, we append separate multi-label heads to the end of the model, with each head corresponding to one of the facets: Materials, Object Types, Disciplines, and Subjects. This design allows the model to simultaneously predict multiple labels across different facets, capturing the diverse visual and contextual characteristics of cultural heritage assets represented in our dataset. By evaluating these baselines, we aim to establish a basis for assessing the performance of image tagging algorithms in the GLAM domain and to provide insights into the challenges and opportunities inherent in automatic metadata annotation within this domain.

The rest of this paper is organized as follows. Section~\ref{sec:related-work} provides an overview of related work, particularly focusing on research related to hierarchical image classification. In Section~\ref{sec:Dataset}, we present a detailed description of the EUFCC-340K dataset, including its collection process, annotation framework, and characteristics. We discuss the design of the baseline models in Section~\ref{sec:method}, and also describe our architecture and implementation. In Section~\ref{sec:experiments} we report on our experimental setup and results, evaluating the performance of the proposed baselines on our dataset, and proposing an assistant tool for GLAM cataloguers. We conclude in  Section~\ref{sec:conclusion} with a summary of our findings and discussion of possible future research.

\section{Related Work}
\label{sec:related-work}
TinyImages~\cite{torralba200880} was one of the earliest attempts to create a vast and varied hierarchical dataset for computer vision, containing 80 million 32x32 low-resolution images collected from the internet by querying every word in the WordNet~\cite{fellbaum1998wordnet} hierarchy. However, the dataset was limited by its low resolution and the noise inherent in its uncurated collection process. Following TinyImages, the ImageNet~\cite{deng2009imagenet, russakovsky2015imagenet} dataset was also organized according to the WordNet hierarchy, offering over 14 million manually verified and high-resolution images categorized into more than 20,000 synsets (WordNet nodes). The dataset provides image-level annotation of a binary label indicating the presence or absence of an object class in the image. A smaller dataset for benchmarking computer vision hierarchical classification models was proposed by Seo \etal \cite{seo2019hierarchical}, by restructuring the Fashion-MNIST dataset~\cite{xiao2017/online} into a simple three-level balanced hierarchy to classify fashion items. 

The large-scale Open Images~\cite{kuznetsova2020open} dataset for object detection provides 15M bounding box annotations that are created according to a semantic hierarchy of classes. The evaluation protocol measures Average Precision (AP) for leaf classes in the hierarchy as normally, while the AP for non-leaf classes AP is computed involving all its ground-truth object instances and all instances of its subclasses.

Hierarchical data also emerges inherently in more specific domains, such as animal or plant classification. The iNaturalist dataset~\cite{van2018inaturalist, van2021benchmarking} for large-scale species classification also offers a hierarchical organization, albeit with a simpler two-level hierarchy consisting of superclasses (``Insects'', ``Fungi'', ``Plants'', etc.) and finer classes. On the other hand, the Herbarium 2021 Half–Earth Challenge Dataset~\cite{de2022herbarium} for species recognition features a challenging real–world image classification task. The dataset includes more than 2.5 million images of vascular plant specimens representing 64,500 taxonomic labels. In addition to labels for species–level and below, labels at higher levels in the taxonomic hierarchy are also included (family and order). 

Diverging from prior works, our EUFCC-340K dataset is specifically designed to meet the unique challenges of Galleries, Libraries, Archives, and Museums (GLAMs) in cataloging cultural heritage assets. The hierarchical structure of the ``Art \& Architecture Thesaurus'' (AAT) offers a domain-specific framework far beyond the capabilities of general-purpose datasets like ImageNet. Moreover, its multi-faceted hierarchy -- spanning Materials, Object Types, Disciplines, and Subjects -- goes beyond object recognition and includes annotations for materials from which objects are made. This not only involves recognizing the object itself but also understanding the materials, which presents a considerably more complex challenge.

\begin{figure}[b]
    \centering
    \includegraphics[width=\textwidth]{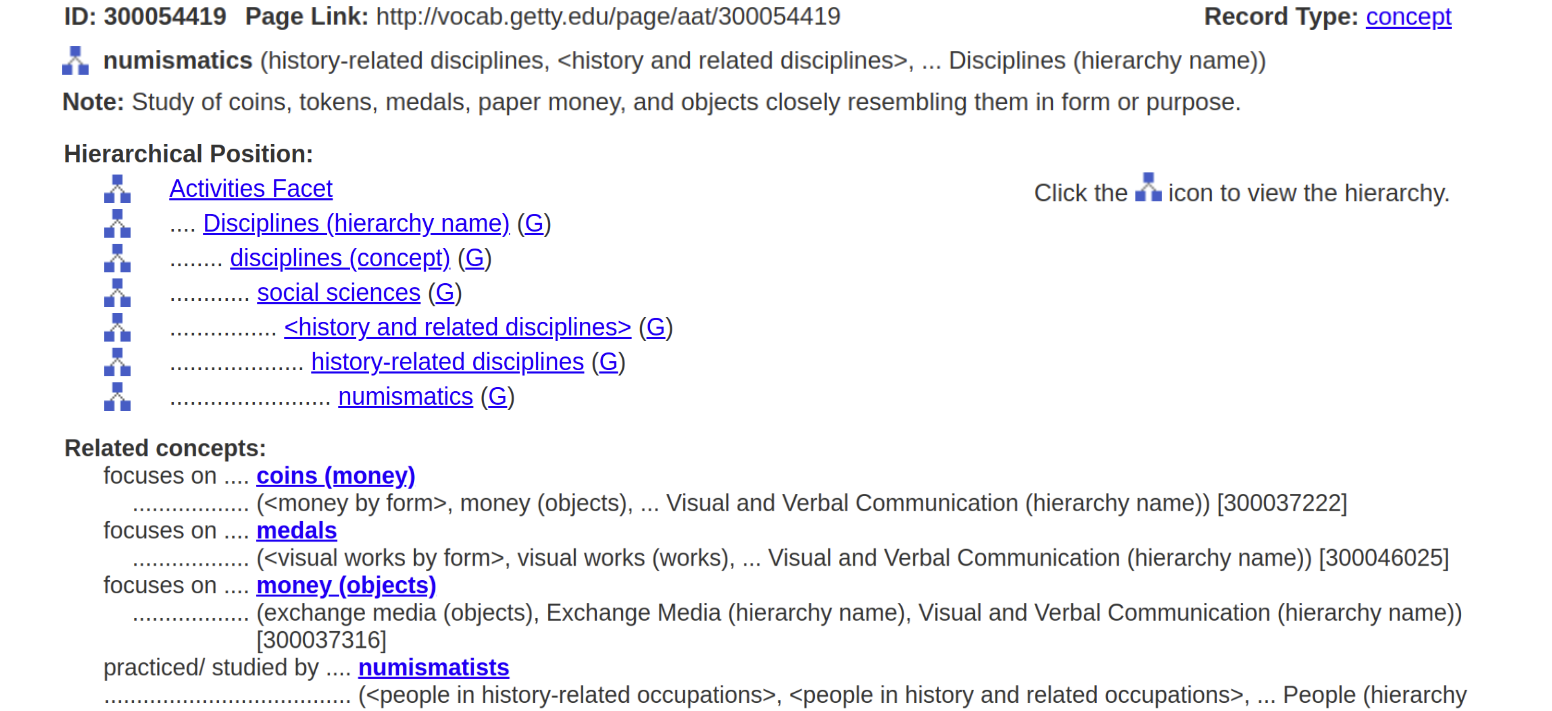}
    \caption{Description and hierarchical position of the AAT term ``Numismatics'' according to the AAT Online search tool.}
    \label{fig:search_getty_results}
\end{figure}

\section{The EUFCC-340K Dataset}
\label{sec:Dataset}
The EUFCC-340K dataset was compiled using the REST API provided by the Europeana portal\footnote{https://pro.europeana.eu/page/apis}, a central repository for cultural heritage collections across Europe. Europeana aggregates a diverse array of cultural artifacts, multimedia content, and traditional records from numerous European institutions, offering a rich bed of resources for academic and research purposes. The platform's comprehensive metadata for each item, detailing various attributes and historical contexts, provides a unique opportunity for in-depth analysis and application in computer vision.

The hierarchical labeling structure of the EUFCC-340K dataset is derived from and aligned with the Getty ``Art \& Architecture Thesaurus (AAT)'' with the help of domain experts. The Getty AAT is a continually evolving vocabulary, designed to support the cataloging, research, and documentation of art, architecture, and other cultural material. Its structured hierarchy enables the detailed classification of concepts and objects and reflects the complex relationships and attributes inherent in the cultural heritage domain. Figure~\ref{fig:search_getty_results} illustrates the description and hierarchical position of the AAT term ``Numismatics'' according to the AAT Online search tool.

    \minisection{Keyword Searches and Filtering.} Initial data collection involved performing keyword searches for the broad ``Collections'' categories, such as ``Furniture'', ``Tool'', ``Costume'', among others. Results were filtered to include only entries with available thumbnails and tagged with ``Reusability: Open,'' ensuring that the dataset comprises images suitable for open research and application. We intentionally excluded categories such as ``Photography'' and ``Painting'' from the primary dataset to avoid the complexities introduced by the overlap of content and form descriptions in artistic representations. Such items were considered separately to maintain the dataset's focus.

    \minisection{Mapping Europeana Concepts to Getty AAT.} For each item retrieved, we extracted the ``concepts'' aligned with the AAT thesaurus and also those assigned by Europeana, that have already a declared/explicit alignment with AAT terms. The Getty AAT's hierarchical structure provided a framework for organizing the dataset's labels, facilitating the representation of complex relationships and attributes inherent in the cultural heritage objects. We created mappings between Europeana concepts and those of the Getty AAT, categorizing each item under four facets (top nodes in AAT) -- Materials, Object Types, Disciplines, and Subjects -- based on concept hierarchy (or hierarchical inheritance) and the criteria of the domain experts. This step was critical in translating Europeana's diverse and rich metadata into a structured format that aligns with the Getty AAT's vocabulary.

    \minisection{Manual Curation and Quality Assurance.} Despite the automated processes involved in data collection and mapping, manual curation plays an important role in ensuring the dataset's quality. Domain experts contributed to cleaning the data, addressing ambiguities and errors that automated processes might overlook. This manual intervention, although challenging, was crucial for maintaining the dataset's integrity and relevance to the GLAM domain. Still, we acknowledge that the dataset might contain noisy annotations in specific categories, as the curation process was not exhaustive at the record level.

    \minisection{Data Provider Information.} Each record was also annotated with information about the data provider, which, while not always directly corresponding to a single museum, offers insights into the source institution. 

\begin{figure}
    \centering
    \includegraphics[width=\textwidth]{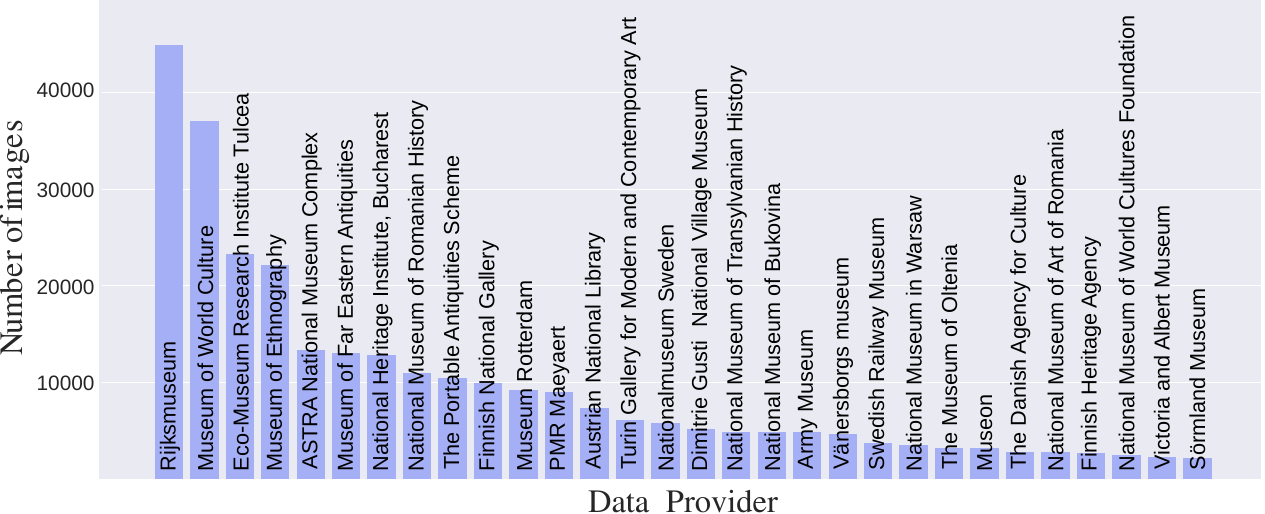}
    \caption{Items' distribution for the $30$ most frequent data providers within the EUFCC-340K dataset.}
    \label{fig:museums_europeana}
\end{figure}

In total, our dataset consists of $346,324$ annotated images gathered from a total of 358 data providers. The contribution from these providers varies significantly, with some contributing as few as a single item, while others offer extensive inventories, including up to $40,000$ items. Figure \ref{fig:museums_europeana}, shows the items' distribution for the $30$ most frequent data providers within the dataset. 

\begin{figure}
  \centering
  \begin{subfigure}{0.47\textwidth}
    \includegraphics[width=\linewidth]{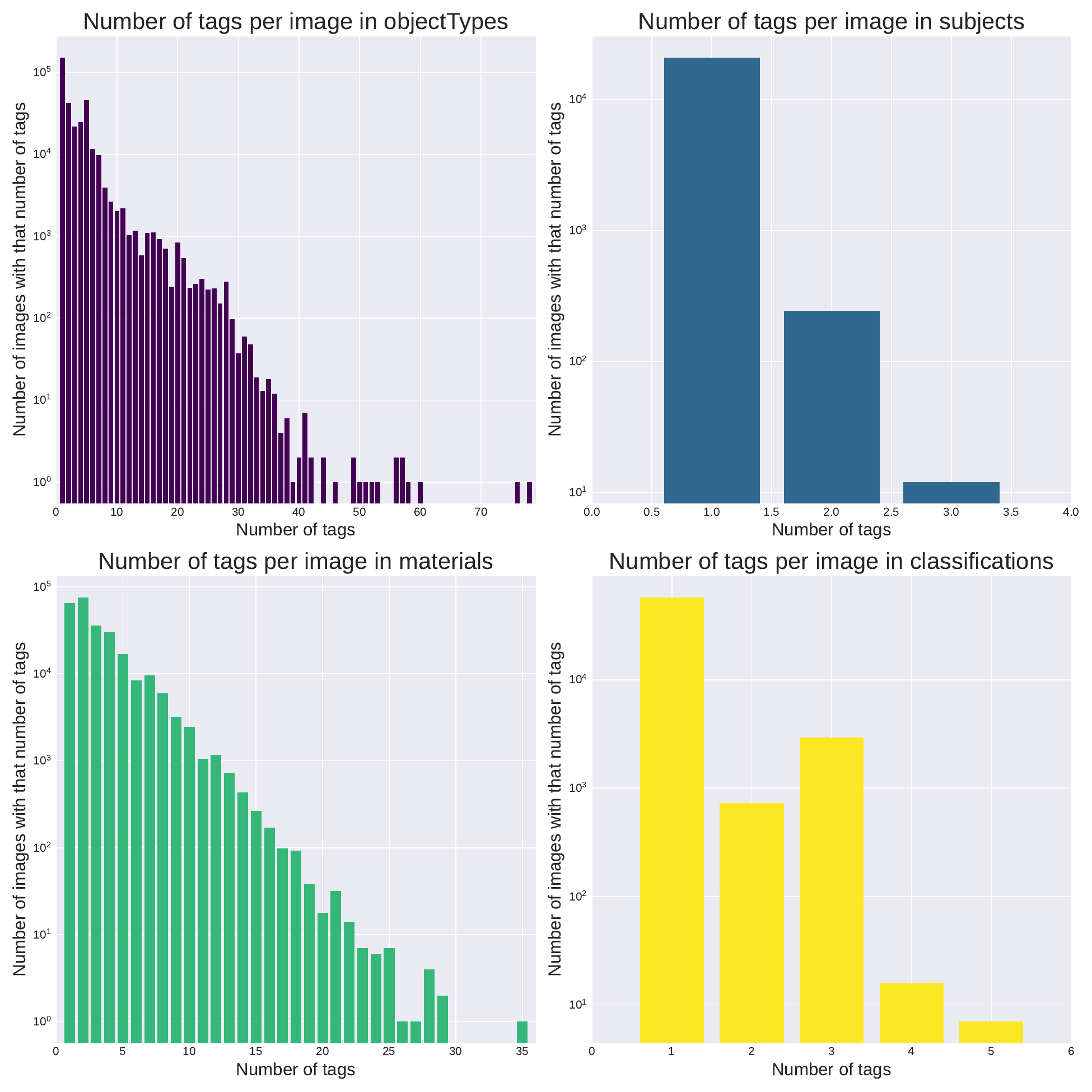} 
    \caption{Number of tags per image}
  \end{subfigure}
  \hfill
  \begin{subfigure}{0.50\textwidth}
    \includegraphics[width=\linewidth]{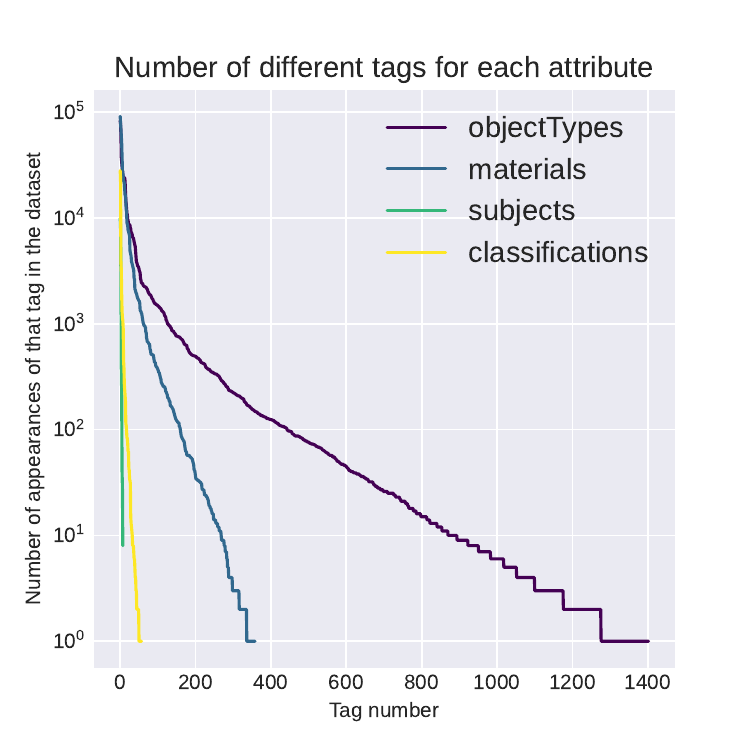} 
    \caption{Number of instances of each tag}
  \end{subfigure}
  \hfill
  \caption{EUFCC-340K statistics: tag count and tag length variation per image.}
  \label{fig:europeana_statistics}
\end{figure}

Figure \ref{fig:europeana_statistics} illustrates the number of tags (class labels) per image and the overall frequency distribution of tags. Based on insights from this analysis, we introduced an additional filter to ensure a minimum of 10 images per tag. Furthermore, we excluded a specific museum,  ``Mobilier National,'' from the dataset due to an excessive number of tags per image, some of which were inaccurately labeled.

The number of unique tags for each facet is 33, 910, 8, and 276 for the ``Classifications'', ``Object Types'', ``Subjects'', and  ``Materials'' respectively. It is worth noting the dataset items' are sometimes partially annotated, \ie not all images contain labels for all facets. Concretely, we have $78,948$ image annotations with one facet, $215,196$ with two facets, $51,734$ with three facets, and $446$ with four facets. Moreover, in some instances, tag labels are assigned only up to a specific node in the hierarchy without necessarily reaching a leaf node. This reflects a practical and realistic use case, where not every image needs to be labeled down to the finest level of detail. Figures \ref{image:europeana} and \ref{image:europeana_labeled} illustrate several examples of the EUFCC-340K dataset.

\include{tikz_samples2}

\begin{figure}
    \centering
    \includegraphics[width=\textwidth]{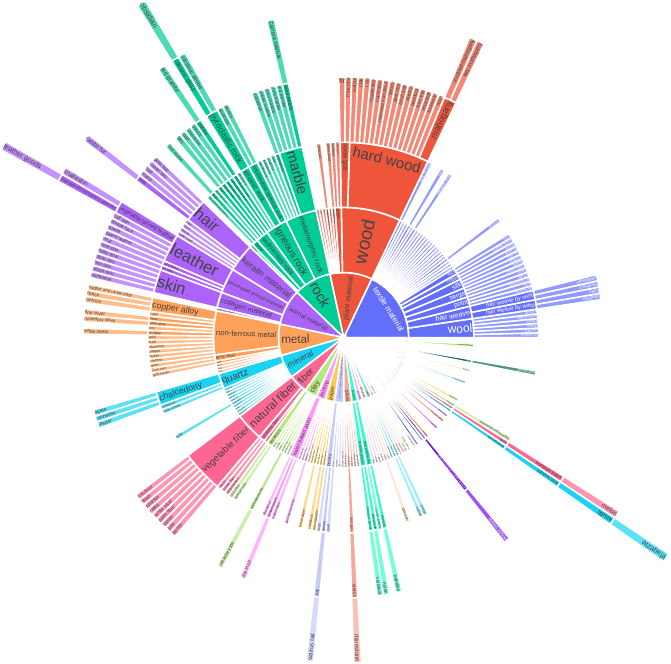}
    \caption{Sunburst chart diagram visualization of the ``Materials'' facet hierarchy. The circle in the centre represents the root node, with the hierarchy expanding outwards from the center. It shows how the outer rings (child nodes) relate to the inner rings (parent nodes in the hierarchy), and how data is distributed across nodes.}
    \label{fig:materials}
\end{figure}

To further illustrate the complex structure of our dataset, Figure \ref{fig:materials} provides a sunburst chart diagram visualizing the ``Materials'' facet hierarchy. This representation highlights the semantic richness and the multi-level nature of some hierarchies within our dataset. The diagram clearly shows the relationship between different levels of the hierarchy, from the central root node outwards to the child nodes, underscoring how data is distributed across various levels. This visualization not only aids in understanding the hierarchical relationships but also showcases the depth and breadth of possible annotations within the dataset.

\minisection{Dataset Splits.} Within the dataset, we have designated two distinct subsets to serve as the test data, collectively representing approximately 10\% of the EUFCC-340K dataset. One subset is identified as the \textit{Outter} test set, and the other as the \textit{Inner} test set. The Outer test set aims to challenge the model with data that may differ significantly from the majority of the training dataset. For this set, we have carefully selected two data providers that account for 2\% to 5\% of the total data each. The first data provider in the Outer Test Set is chosen based on maximizing the Frechet Inception Distance (FID)~\cite{parmar2022aliased} with the entire dataset while ensuring that the image types are diverse and not exclusively from a single category (e.g., stamps). The second data provider in the Outer test set was selected as the one with FID to the first one. The data providers conforming to the Outer test set are: ``The Portable Antiquities Scheme'' and ``PMR Maeyaert''.

\begin{figure}
    \centering
    \includegraphics[width=\textwidth]{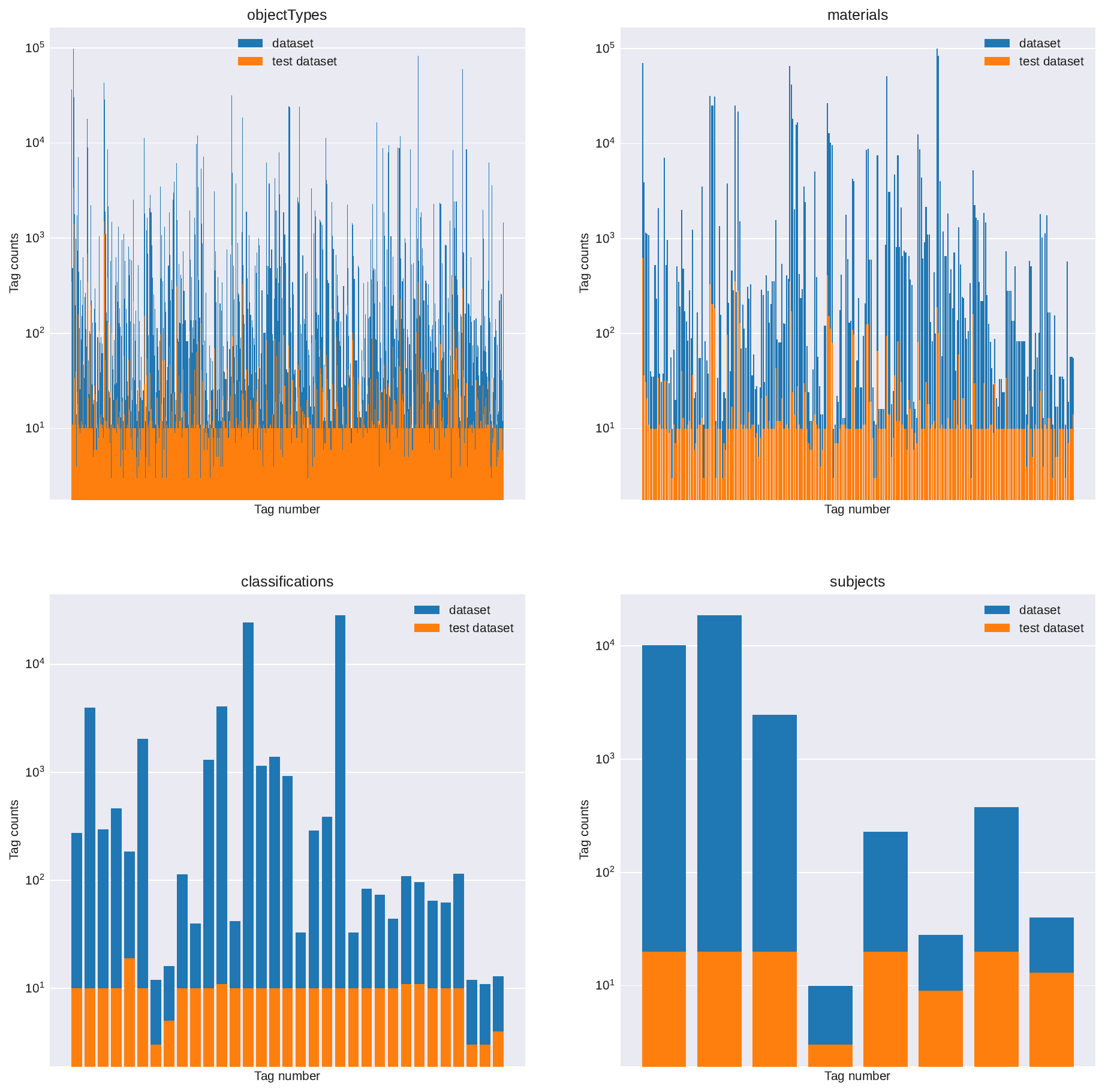}
    \caption{EUFC-340K dataset split for the training set and Inner test set.}
    \label{fig:dataset_partition}
\end{figure}

For the Inner Test Set, we adhered to a balanced selection criterion, aiming for a 33\% allocation of images for testing, with minimum thresholds set for each tag category (at least 5 images for classifications, object types, and materials, and 20 for subjects). This strategy ensures a diverse and representative sample of the dataset’s breadth. Figure \ref{fig:dataset_partition} illustrates the split of the training and the Inner Test Set, highlighting the methodological constraints applied to achieve balance and diversity. Although our dataset features tags with varying frequencies, our split strategy occasionally results in minor data distribution peaks, as depicted in the figure. The validation set, constituting another 10\% of the dataset, is derived using similar criteria, with the remainder allocated for training. The whole dataset and official splits are made publicly available\footnote{\url{https://github.com/cesc47/EUFCC-340K}.}.

\section{Baseline Methods}
\label{sec:method}

In this section, we introduce the baseline models that we use to benchmark performance on the proposed EUFCC-340K dataset for image tagging in the GLAM domain. We employ two distinct types of baseline models: vision-only models that leverage a state-of-the-art CNN architecture for image classification, and a state-of-the-art multi-modal model that projects visual and textual modalities into a shared common feature space. For each model, we detail their architecture and training strategies in the following subsections.

\subsection{CNN-based vision models}

Our vision-only baselines extend a state-of-the-art image classification CNN backbone, the ConvNeXT$_{base}$ model \cite{liu2022convnet}, by incorporating multiple classification heads. Each head is dedicated to one of the EUFCC-340K facets: Materials, Object Types, Disciplines, and Subjects. This multi-head setup enables the model to perform multi-label classification across various facets simultaneously. During training, we compute a separate loss for each classification head, which allows specialized tuning of the model's ability to distinguish between class labels within each facet. This section details the various approaches and loss calculation methods used for each baseline within the vision-only architecture.

\begin{figure}
    \centering
    \includegraphics[width=\textwidth]{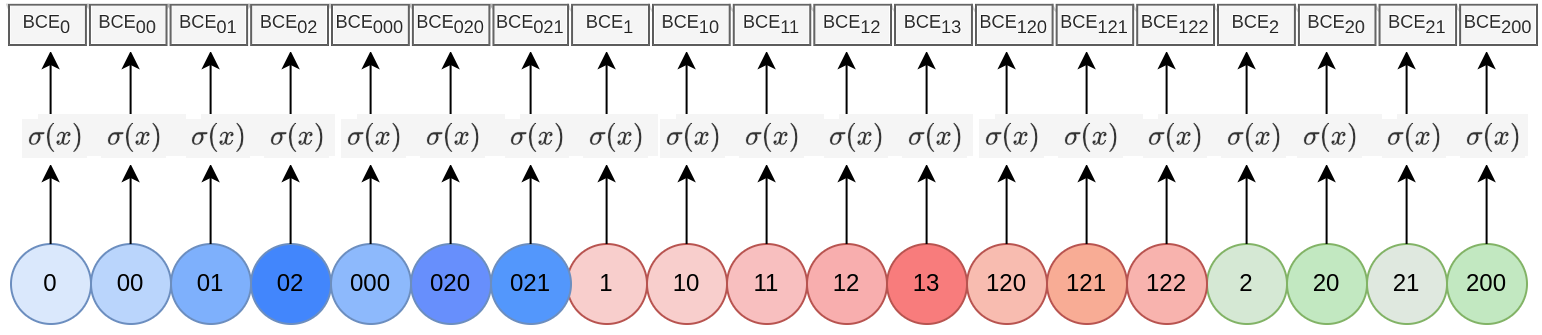}
    \caption{An illustrative example depicting a multi-label classification head. Each output neuron behaves as a binary classifier for one label within a given hierarchy where the node labeled as ``0'' is the parent of nodes \{``00'', ``01'',  ``02''\}, and so on.}
\label{fig:multi_label_example}
\end{figure}

\subsubsection{Multi-label classification}
In this baseline, each output neuron in each of the heads has an individual sigmoid activation function ($\sigma(x) = \frac{1}{1 + e^{-x}}
$) that can be independently activated, allowing the model to predict the presence or absence of each tag individually. In other words, each output is a binary classifier that predicts whether each tag is activated (present) or not for the input image. Figure~\ref{fig:multi_label_example} illustrates the outputs of a multi-label head with 19 outputs, each output corresponds to a node in a given tag hierarchy with three levels, such as the node labeled as ``0'' is the parent of nodes \{``00'', ``01'', ``02''\}, and so on. The network is trained with Binary Cross Entropy (BCE) loss: $l(y, \hat{y}) = y \cdot \log(\hat{y}) + (1 - y) \cdot \log(1-\hat{y})$. \\

\subsubsection{Softmax classification}
\begin{figure}
    \centering
    \includegraphics[width=\textwidth]{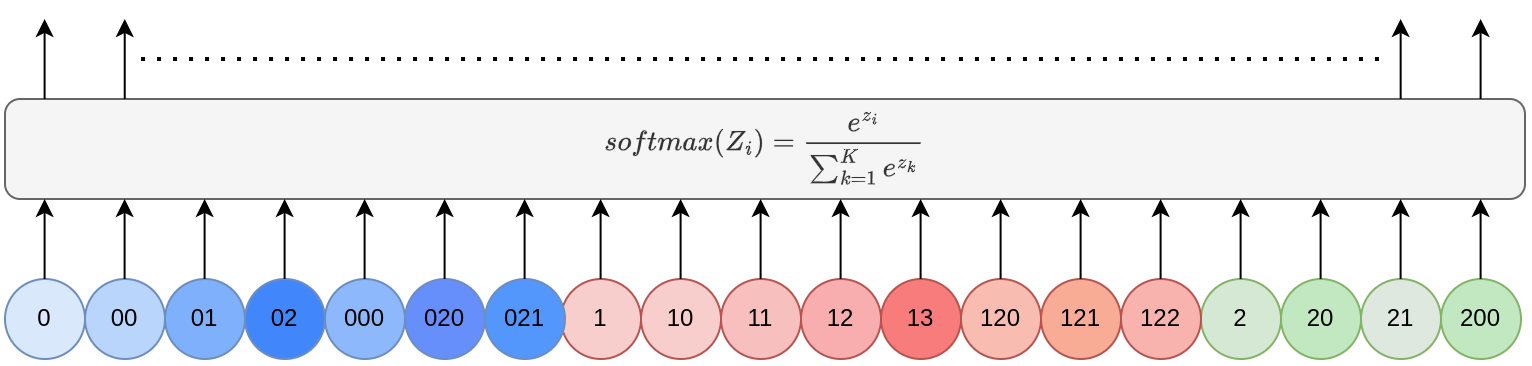}
    \caption{Softmax classification head. The softmax output can be interpreted as a probability distribution over classes.}
\label{fig:softmax_head}
\end{figure}

In this baseline, we follow prior works~\cite{sun2017revisiting,joulin2016learning}
that suggest that softmax classification can be very effective in multi-label settings with large numbers of classes. Softmax classification heads, as illustrated in Figure~\ref{fig:softmax_head}, are the default choice for multi-class classification problems (where a given input belongs only to one single class), as the softmax output can be interpreted as a probability distribution over classes. In our context, the use of a softmax head in a multi-label setting involves the random selection of one tag from each attribute during the training phase and applying the cross-entropy loss function: 
$ H(Y, \hat{Y}) = -\sum_{i=1}^{N} Y_i \log(\hat{Y}_i)$. 
At inference time, the predicted tags are sorted in descending order of probability.

\subsubsection{Hierarchical Softmax} 
\begin{figure}
    \centering
    \includegraphics[width=\textwidth]{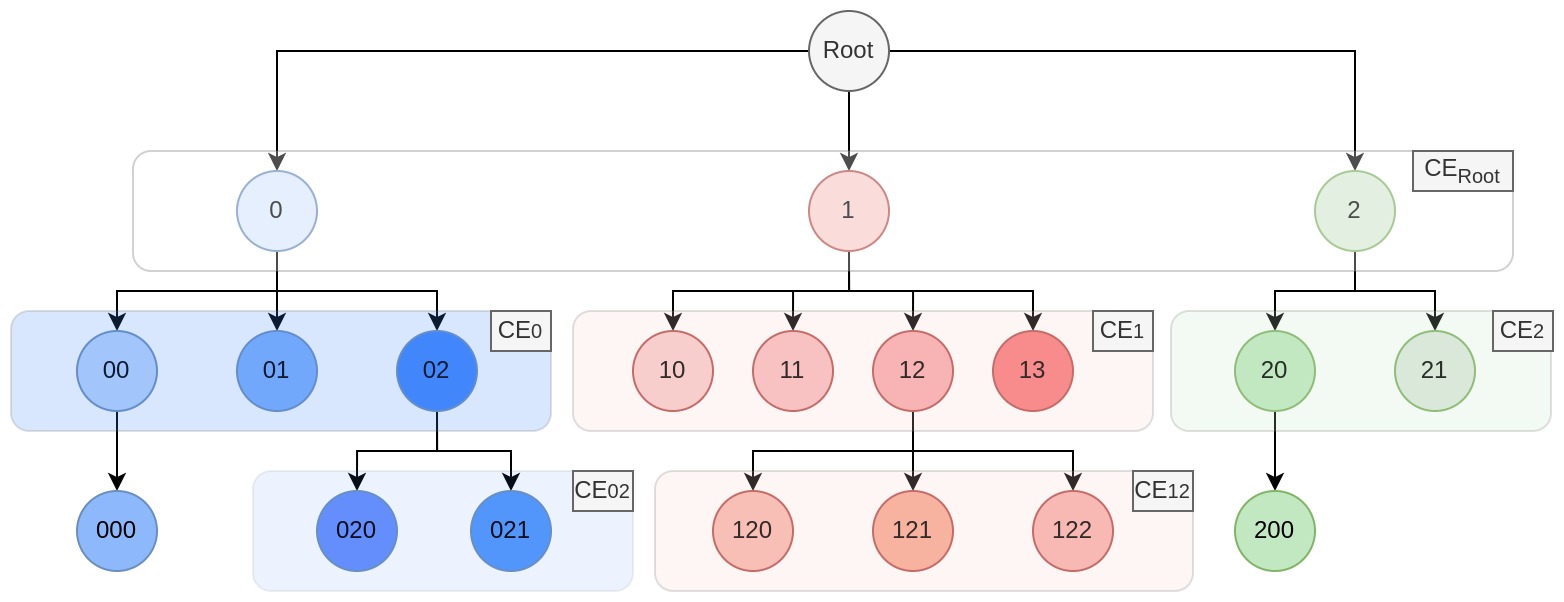}
    \caption{An illustrative tree representation featuring a hierarchical softmax head, highlighting the specific nodes where cross-entropy (CE) losses might be computed.}
    \label{fig:tree}
\end{figure}

\begin{figure}[b]
    \centering
    \includegraphics[width=\textwidth]{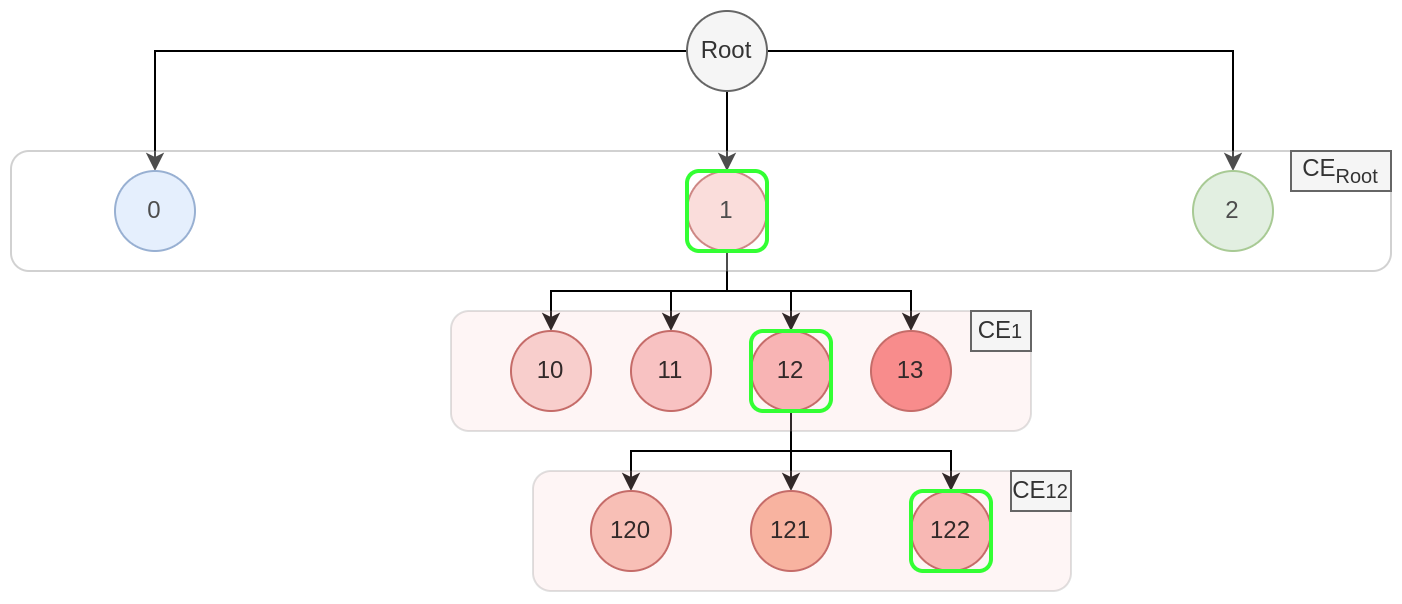}
    \caption{An illustrative tree diagram demonstrating the computation of the loss function of a given input that is annotated with the term ``$122$'' in its ground-truth.}
    \label{fig:tree_example}
\end{figure}

The baseline models described thus far do not account for the hierarchy among class labels. They simplify (flatten) the class hierarchy, treating each output neuron either as an independent classifier in a multi-label setup or as part of a multi-class model that predicts a probability distribution over classes using a softmax output. In contrast, the Hierarchical Softmax baseline incorporates the class hierarchy into the labeling process. This approach provides context and structure, enabling the model to label images with multiple levels of granularity. We follow prior works~\cite{redmon2017yolo9000} with the difference that we have several tagging heads instead of one. To perform classification with one of the hierarchical facets of the EUFCC-340K dataset we
predict conditional probabilities at every node for the probability of each child node of a parent node given that parent node. As illustrated in Figure~\ref{fig:tree} training this model involves computing a cross-entropy loss at every node that is relevant for a given input. For example, if a given input has the term ``$122$'' in its ground truth annotations the losses $CE_{Root}$, $CE_{1}$, and $CE_{12}$ must be calculated and backpropagated, as depicted in figure \ref{fig:tree_example}. 

Our hierarchical softmax models make use of weighted cross-entropy losses (assigning higher weights to minority classes and lower weights to majority classes) to facilitate unbiased training. Furthermore, in the experimental section, we also evaluate a modified version of the hierarchical softmax loss, where the loss is calculated per level instead of at individual tree nodes. This approach is adopted to prevent an emphasis on predictions that consistently follow a single-label path and to encourage a more diverse set of predictions.

\subsection{Multi-modal baselines}

Our multi-modal baselines involve fine-tuning a pre-trained CLIP model~\cite{radford2021learning} using image-text pairs. The CLIP (Contrastive Language–Image Pre-training) model is designed to understand and generate representations that link textual and visual information, making it ideal for tasks that involve correlating images with descriptive texts. In our context, the text modality comprises all labels of a given image for a given facet, posing the challenge of effectively integrating them into a single text prompt. To address this, we explore diverse strategies for constructing text prompts, including: (1) using all raw tags separated by commas, (2) using a randomly selected single tag per prompt (similar to our approach in the softmax baseline), and (3) experimenting with template prompts tailored to distinct attributes by integrating relevant tags for each image. For enhanced clarity, consider the ``Materials'' facet, where the ground truth annotations for a given image might be  ``\textit{metal}'', ``\textit{non-ferrous metal}'', and ``\textit{bronze}''. The paired text for this image, depending on the chosen strategy, could be: (1) ``{\fontfamily{lmtt}\selectfont metal, non-ferrous metal, bronze}'', (2) one of ``{\fontfamily{lmtt}\selectfont metal}'', ``{\fontfamily{lmtt}\selectfont non-ferrous metal}'', or ``{\fontfamily{lmtt}\selectfont bronze}'', or (3) ``{\fontfamily{lmtt}\selectfont an image of an object made of <TAG>}'' where {\fontfamily{lmtt}\selectfont <TAG>} is a placeholder that can be substituted by any of the tags in the ground-truth. In the next section, we will compare the results of several CLIP baselines fine-tuned using all these prompt engineering strategies.


\section{Experiments}
\label{sec:experiments}
In this section, we first present the proposed evaluation framework for the EUFCC-340K dataset, then we give the implementation details for training all baseline models discussed in Section~\ref{sec:method}. Finally, we present and discuss the results obtained with all evaluated models, and present a tool designed to assist human expert annotators. 

\subsection{Evaluation Metrics} 
\label{sec:metrics}
Image tagging models are evaluated based on their ability to predict, for a given image, the relevance of each tag in the vocabulary, akin to creating a list of tags sorted by their predicted relevance. Then the model performance is measured using information retrieval metrics such as mean Average Precision (mAP). While mAP is widely used due to its ability to capture both precision and recall, it becomes less effective in datasets like EUFCC-430K where there is a high variability in the number of relevant tags across different images (see figure~\ref{fig:europeana_statistics}). Such variability can lead to inconsistent evaluations, as mAP may disproportionately favor queries with fewer relevant items or penalize those with more, due to its integrated recall component. To provide a more stable and meaningful evaluation, we have selected alternative metrics better suited to our dataset. We use R-Precision, Accuracy@K, and Average Rank Position, which focus on the precise assessment of how accurately tags are assigned without being affected by the variable number of tags per image. These metrics provide clear insights into both the precision of the tag assignments and the overall ranking quality of the model’s outputs.

\noindent
\textbf{R-Precision} assesses image tagging effectiveness by considering the precision of the top $R$ tags, where $R$ is the number of relevant tags known for a specific image. Formally:
\begin{equation}
\text{R-Precision} = \frac{1}{N} \sum_{i=1}^{N} \frac{r_i}{R_i},
\end{equation}
where $N$ is the total number of images in the test set, $R_i$ is the number of relevant tags for the i-th image according to the ground-truth, and $r_i$ is the number of correct tags among the top $R_i$ tags returned by the system. Note that this measure is equivalent to both precision and recall at the $R_i^{th}$ position.
In practical terms, if there are $R$ tags relevant to an image, R-Precision looks at the top $R$ tags returned by the system, counts how many of them are relevant ($r$), and calculates the ratio. 

\minisection{Accuracy@k} assesses the performance of the model by measuring how often at least one of the ground-
truth tags appears in the $k$ highest-ranked predictions:
\begin{equation}
\text{Acc@k} = \frac{1}{N} \sum_{i=1}^{N} \mathbbm{1} [ \text{GT}_i \cap \text{top-K}_i],
\end{equation}
where $\mathbbm{1} [\text{GT}_i \cap \text{top-K}_i]$ takes the value $1$ or $0$ depending if at least one of the ground-truth tags (GT) is present in the top-K model predictions or not. In our experiments, we evaluate accuracy at $k=1$ and $k=10$.

\minisection{Average Rank Position}
determines the average position of the ground truth labels in the output list, providing a quantitative measure of how well the model's predictions align with the true ranking. A lower AvgRankPos value indicates better performance. Formally,  AvgRankPos is defined as follows:
\begin{equation}
\text{AvgRankPos} = \frac{1}{N} \sum_{i=1}^{N} \sum_{j=1}^{R_i} \frac{Pos_{i,j}}{R_i},
\end{equation}
where  $R_i$ is the number of relevant tags for the i-th image according to the ground-truth, and $Pos_{i,j}$ is the ranking position of j-th tag for the i-th image. 

\subsection{Implementation details}
\minisection{Vision models.} All baselines were trained using the default pre-trained weights (from ImageNet) with four NVIDIA A40 GPUs, with a batch size of 128, and optimized using the AdamW optimizer, weight decay was set to 0.05, and a learning rate of 1e-4 during 25 epochs. The learning rate was scheduled using the cosine annealing strategy. The training process incorporated simple data augmentations: a random resized crop of 224 and a random horizontal flip. In addition, repeated augmentation sampling\footnote{\url{https://github.com/facebookresearch/deit/blob/main/samplers.py}} (RA-Sampler) was used. Each training lasted 12 hours approximately.

\minisection{Multi-modal models.} All baselines were trained with four NVIDIA A40 GPUs, with a batch size of 64, during 100 epochs, and utilizing a learning rate of $3 \times 10^{-5}$. Beta1 and beta2 parameters of Adam optimizer were configured at 0.9 and 0.99, respectively, accompanied by an epsilon value of $10^{-8}$ and a weight decay of 0.1. Other specifications included a warm-up period of 500 steps, gradient clipping with a norm of 1, and the adoption of the ViT-B-16 model architecture pre-trained on the WIT dataset~\cite{radford2021learning}. The duration of fine-tuning varied based on the model's design—ranging from 3 to 7 days approximately—depending on whether it divided tags (resulting in a slower training process with more instances) or worked with all tags from an image collectively.

\subsection{Automatic metadata annotation results}

Tables~\ref{tab:test_id} and \ref{tab:test_ood} present the results of all baseline models for the Inner and Outer test sets respectively. We provide results for all baselines defined in section~\ref{sec:method} -- Multi-label, Softmax, Hierarchical Softmax (H-Softmax), and CLIP models with different prompting strategies (all tags, one tag, and one tag with template prompt) -- using all metrics described in section~\ref{sec:metrics}. For the Multi-label baselines, we provide also results for a multi-label model trained with a single head (``Materials'' facet). For the Softmax baselines, we provide two variants: one trained with cross-entropy loss (Softmax), and another with weighted cross-entropy loss (WCE). For the hierarchical softmax models, we provide results of two variants in which losses are calculated at the node level and the tree depth level respectively. For the CLIP model, we also provide results in a zero-shot setting, without any fine-tuning. Finally, we provide results for two ensembles that combine the outputs of the multi-modal (ML), softmax (S), and CLIP models.

\begin{table}
\centering
\begin{adjustbox}{width=\textwidth}
\setlength{\tabcolsep}{1.1pt}.

\begin{tabular}{l|llll|llll|llll|llll}
\toprule
\textbf{Method}             & \multicolumn{4}{c|}{\textbf{R-Precision}}                                                                  & \multicolumn{4}{c|}{\textbf{Acc@1}}                                                                        & \multicolumn{4}{c|}{\textbf{Acc@10}}                                                                       & \multicolumn{4}{c}{\textbf{AvgRankPos}}                                                                  \\
\multicolumn{1}{l|}{}       & \multicolumn{1}{c|}{OT}  & \multicolumn{1}{c|}{M}   & \multicolumn{1}{c|}{C}   & S                         & \multicolumn{1}{c|}{OT}  & \multicolumn{1}{c|}{M}   & \multicolumn{1}{c|}{C}   & S                         & \multicolumn{1}{c|}{OT}  & \multicolumn{1}{c|}{M}   & \multicolumn{1}{c|}{C}   & S                         & \multicolumn{1}{c|}{OT}   & \multicolumn{1}{c|}{M}   & \multicolumn{1}{c|}{C}  & S                       \\ \midrule
Multi-label                 & {\ul 0.60}               & 0.72                     & 0.88                     & 0.82                      & 0.78                     & 0.83                     & 0.88                     & 0.83                      & {\ul 0.97}               & {\ul 0.98}               & {\ul 0.99}               & 1.00                      & {\ul 11.8}                & 4.4                      & {\ul 1.4}               & 1.4                     \\
Multi-label (M)        & -                        & 0.72                     & -                        & -                         & -                        & 0.83                     & -                        & -                         & -                        & 0.97                     & -                        & -                         & -                         & 5.3                      & -                       & -                       \\ \midrule
Softmax                     & 0.59                     & {\ul 0.75}               & \textbf{0.92}            & 0.83                      & \textbf{0.86}            & {\ul 0.89}               & \textbf{0.93}            & 0.85                      & \textbf{0.98}            & \textbf{0.99}            & {\ul 0.99}               & 1.00                      & 15.0                      & 4.0                      & {\ul 1.4}               & 1.3                     \\
Softmax (WCE)               & 0.39                     & 0.46                     & 0.84                     & 0.83                      & 0.66                     & 0.59                     & 0.85                     & 0.86                      & 0.93                     & 0.92                     & {\ul 0.99}               & 1.00                      & 43.1                      & 12.1                     & 1.8                     & 1.3                     \\
H-Softmax (levels)          & 0.46                     & 0.64                     & 0.85                     & 0.78                      & 0.84                     & 0.88                     & 0.86                     & 0.80                      & \textbf{0.98}            & \textbf{0.99}            & {\ul 0.99}               & 1.00                      & 81.2                      & 13.6                     & 2.2                     & 1.4                     \\
H-Softmax (nodes)           & 0.55                     & 0.72                     & 0.84                     & 0.82                      & 0.78                     & \textbf{0.90}            & 0.86                     & 0.84                      & \textbf{0.98}            & \textbf{0.99}            & {\ul 0.99}               & 1.00                      & 36.4                      & 5.5                      & 2.1                     & 1.3                     \\ \midrule
Ensemble: ML+S              & \textbf{0.66}            & \textbf{0.79}            & {\ul 0.91}               & 0.86                      & \textbf{0.86}            & {\ul 0.89}               & {\ul 0.92}               & 0.86                      & \textbf{0.98}            & \textbf{0.99}            & \textbf{1.00}            & 1.00                      & \textbf{9.7}              & \textbf{3.1}             & \textbf{1.3}            & 1.2                     \\ \midrule
CLIP zero-shot              & \multicolumn{1}{l}{0.07} & \multicolumn{1}{l}{0.06} & \multicolumn{1}{l}{0.42} & \multicolumn{1}{l|}{0.86} & \multicolumn{1}{l}{0.10} & \multicolumn{1}{l}{0.07} & \multicolumn{1}{l}{0.46} & \multicolumn{1}{l|}{0.86} & \multicolumn{1}{l}{0.44} & \multicolumn{1}{l}{0.43} & \multicolumn{1}{l}{0.77} & \multicolumn{1}{l|}{1.00} & \multicolumn{1}{l}{199.6} & \multicolumn{1}{l}{58.5} & \multicolumn{1}{l}{7.3} & \multicolumn{1}{l}{{\ul 1.1}} \\
CLIP 1 tag   & 0.51                     & 0.71                     & 0.48                     & {\ul 0.88}                & 0.68                     & 0.81                     & 0.46                     & {\ul 0.88}                & 0.85                     & 0.90                     & {\ul \textbf{1.00}}      & 1.00                      & 37.8                      & 10.8                     & 2.9               & {\ul 1.1}                     \\
CLIP all tags             & 0.58                     & 0.15                     & \textbf{0.92}            & 0.78                      & 0.59                     & 0.30                     & {\ul 0.92}               & 0.78                      & 0.88                     & 0.65                     & {\ul 0.99}               & 1.00                      & 50.4                      & 52.1                     & {\ul 1.4}               & 1.6            \\
CLIP 1 tag prompt & 0.52                     & 0.68                     & 0.60                     & \textbf{1.00}             & 0.69                     & 0.76                     & 0.58                     & \textbf{1.00}             & 0.88                     & 0.92                     & 0.92                     & 1.00                      & 41.4                      & 10.5                     & 3.4                     &  \textbf{1.0}      \\ \midrule
Ens: ML+S+CLIP         & \textbf{0.66}            & \textbf{0.79}            & {\ul 0.91}               & 0.83                      & {\ul 0.85}               & {\ul 0.89}               & 0.91                     & 0.83                      & {\ul 0.97}               & \textbf{0.99}            & \textbf{1.00}            & 1.00                      & 13.8                      & {\ul 3.5}                & \textbf{1.3}            & 1.3 \\
\bottomrule
\end{tabular}
\end{adjustbox}
\vspace{1mm}
\caption{Inner test set results for automatic metadata annotation on the different facets EUFCC-340K. H-Softmax stands for Hierarchical Softmax. \textit{Facet Keys: (M) Materials, (S) Subjects, (OT) Object Type, (C) Classifications}.}
\label{tab:test_id}
\end{table}

In Table~\ref{tab:test_id} we appreciate several key insights. First, the multi-label baseline with a single head -- Multi-label (M) -- performs almost equally as the multi-head (Multi-label) model on the ``Materials'' facet. This validates our approach of extending the CNN backbone with multiple heads. Second, the Softmax baseline outperforms all other individual vision models, with the Multi-label baseline performing competitively. Third, all fine-tuned multi-modal CLIP models outperform the zero-shot CLIP model. Among them, the CLIP model trained with template prompts has an overall better performance but is below the Softmax baseline in most cases. Finally, we appreciate that the ensemble methods provide an overall extra boost in all facets and metrics as expected. 

\begin{table}
\centering
\begin{adjustbox}{width=\textwidth}
\setlength{\tabcolsep}{1.1pt}.
\begin{tabular}{l|llll|llll|llll|llll}
\toprule
\textbf{Method}             & \multicolumn{4}{c|}{\textbf{R-Precision}}                                                 & \multicolumn{4}{c|}{\textbf{Acc@1}}                                                       & \multicolumn{4}{c|}{\textbf{Acc@10}}                                                                  & \multicolumn{4}{c}{\textbf{AvgRankPos}}                                                  \\
                            & \multicolumn{1}{c|}{OT} & \multicolumn{1}{c|}{M} & \multicolumn{1}{c|}{C} & S             & \multicolumn{1}{c|}{OT} & \multicolumn{1}{c|}{M} & \multicolumn{1}{c|}{C} & S             & \multicolumn{1}{c|}{OT} & \multicolumn{1}{c|}{M} & \multicolumn{1}{c|}{C} & \multicolumn{1}{c|}{S}    & \multicolumn{1}{c|}{OT} & \multicolumn{1}{c|}{M} & \multicolumn{1}{c|}{C} & S            \\ \midrule
Multi-label                 & 0.45                    & \textbf{0.85}          & 0.48                   & {\ul 0.91}    & 0.47                    & \textbf{0.92}          & 0.49                   & {\ul 0.91}    & 0.74                    & {\ul 0.98}             & 0.84                   & \multicolumn{1}{c|}{1.00} & 101.8                   & {\ul 2.3}              & 5.3                    & {\ul 1.2}    \\
Multi-label (M)        & -                       & {\ul 0.84}             & -                      & -             & -                       & {\ul 0.91}             & -                      & -             & -                       & {\ul 0.98}             & -                      & \multicolumn{1}{c|}{-}    & -                       & 2.5                    & -                      & -            \\ \midrule
Softmax                     & 0.50                    & 0.83                   & 0.47                   & 0.87          & 0.51                    & 0.90                   & 0.48                   & 0.87          & 0.74                    & {\ul 0.98}             & 0.76                   & \multicolumn{1}{c|}{1.00} & 93.2                    & 2.4                    & 6.5                    & {\ul 1.2}    \\
Softmax (WCE)               & 0.31                    & 0.61                   & 0.35                   & 0.84          & 0.32                    & 0.60                   & 0.34                   & 0.85          & 0.65                    & 0.93                   & 0.64                   & \multicolumn{1}{c|}{1.00} & 143.8                   & 5.9                    & 11.9                   & {\ul 1.2}    \\
H-Softmax (levels)          & 0.50                    & 0.81                   & 0.49                   & 0.87          & 0.56                    & 0.90                   & 0.49                   & 0.87          & 0.82                    & \textbf{0.99}          & 0.72                   & \multicolumn{1}{c|}{1.00} & 46.9                    & 4.1                    & 7.3                    & 1.3          \\
H-Softmax (nodes)           & 0.42                    & {\ul 0.84}             & 0.48                   & 0.82          & 0.44                    & 0.90                   & 0.47                   & 0.83          & 0.76                    & 0.97                   & 0.72                   & \multicolumn{1}{c|}{1.00} & 58.5                    & 3.1                    & 7.4                    & 1.3          \\ \midrule
Ensemble: ML+S              & 0.50                    & {\ul 0.84}             & 0.48                   & 0.88          & 0.51                    & 0.90                   & 0.49                   & 0.89          & 0.76                    & \textbf{0.99}          & 0.81                   & \multicolumn{1}{c|}{1.00} & 99.7                    & {\ul 2.3}              & 5.7                    & {\ul 1.2}    \\ \midrule
CLIP zero-shot              & 0.01                    & 0.10                   & 0.01                   & 0.67          & 0.01                    & 0.13                   & 0.01                   & 0.67          & 0.05                    & 0.43                   & 0.37                   & \multicolumn{1}{c|}{1.00} & 511.8                   & 47.8                   & 12.2                   & 1.7          \\
CLIP 1 tag    & \textbf{0.75}           & 0.51                   & \textbf{0.87}          & 0.81          & \textbf{0.79}           & 0.52                   & \textbf{0.88}          & 0.81          & {\ul 0.85}              & 0.78                   & \textbf{0.96}          & \multicolumn{1}{c|}{1.00} & \textbf{19.1}           & 15.8                   & \textbf{2.0}           & 1.4          \\
CLIP all tags              & 0.19                    & 0.28                   & 0.55                   & \textbf{1.00} & 0.33                    & 0.42                   & 0.54                   & \textbf{1.00} & 0.74                    & 0.84                   & 0.89                   & \multicolumn{1}{c|}{1.00} & 146.2                   & 41.5                   & 4.3                    & \textbf{1.0} \\
CLIP 1 tag prompt & {\ul 0.73}              & 0.47                   & {\ul 0.83}             & 0.74          & {\ul 0.77}              & 0.52                   & {\ul 0.83}             & 0.74          & \textbf{0.86}           & 0.83                   & {\ul 0.95}             & \multicolumn{1}{c|}{1.00} & {\ul 19.6}              & 17.1                   & {\ul 2.2}              & 1.4          \\ \midrule
Ens: ML+S+CLIP         & 0.50                    & {\ul 0.84}             & 0.48                   & {\ul 0.91}    & 0.52                    & 0.90                   & 0.48                   & {\ul 0.91}    & 0.76                    & \textbf{0.99}          & 0.93                   & 1.00                      & 44.9                    & \textbf{2.2}           & 4.1                    & {\ul 1.2} \\
\bottomrule
\end{tabular}
\end{adjustbox}
\vspace{1mm}
\caption{Outer test set results for automatic metadata annotation on the different facets EUFCC-340K. H-Softmax stands for Hierarchical Softmax. \textit{Facet Keys: (M) Materials, (S) Subjects, (OT) Object Type, (C) Classifications}.}
\label{tab:test_ood}
\end{table}

In Table~\ref{tab:test_ood} we appreciate that the results on the Outer test set are lower in general than in the Inner set. This was expected, as the Outer test set is designed to challenge the model with data that differs from the training set (see section~\ref{sec:Dataset} for details). In this setting, we appreciate that the results of CLIP models slightly outperform those of the visual models, demonstrating superior generalization. It is worth noting that in certain cases a tag label may be present in the Outer test set even though it was not included during the model's training phase, simply because it did not appear in our training dataset. In such cases, we choose to ignore this tag when calculating the evaluation metrics, as it falls outside the model's scope.

Finally, Table \ref{tab:summarized} summarizes the results of Tables~\ref{tab:test_id} and~\ref{tab:test_ood} by showing the mean value of each metric across all facets. The key insights we extract from this summary are as follows: (1) vision-only models perform better in the Inner test set, while CLIP-based models perform better in the Outer test set; and (2) model ensembles that combine the outputs of both visual-only and multi-modal baselines provide a balanced performance across both test scenarios, at the cost of an increased computation cost.

\begin{table}
\centering
\begin{adjustbox}{width=1\textwidth}
\setlength{\tabcolsep}{1.1pt}.
\begin{tabular}{l|cccc|cccc}
\toprule
\multicolumn{1}{c|}{\textbf{Method}} & \multicolumn{4}{c|}{\textbf{Test ID}}                                                                                                & \multicolumn{4}{c}{\textbf{Test OOD}}                                                                                                \\ 
\multicolumn{1}{c|}{\textbf{}}       & \multicolumn{1}{c|}{\textbf{R-Prec}} & \multicolumn{1}{c|}{\textbf{Acc@1}} & \multicolumn{1}{c|}{\textbf{Acc@10}} & \textbf{AvgRP} & \multicolumn{1}{c|}{\textbf{R-Prec}} & \multicolumn{1}{c|}{\textbf{Acc@1}} & \multicolumn{1}{c|}{\textbf{Acc@10}} & \textbf{AvgRP} \\ \midrule
Multi-label                          & 0.76                              & 0.83                                & \textbf{0.99}                        & {\ul 4.8}           & 0.67                              & 0.70                                & 0.89                                 & 27.7                \\ \midrule
Softmax                              & 0.77                              & \textbf{0.88}                       & \textbf{0.99}                        & 5.4                 & 0.67                              & 0.69                                & 0.87                                 & 25.8                \\
Softmax (WCE)                        & 0.63                              & 0.74                                & {\ul 0.96}                           & 14.6                & 0.53                              & 0.53                                & 0.80                                 & 40.7                \\
H-Softmax (levels)                   & 0.68                              & 0.85                                & \textbf{0.99}                        & 24.6                & 0.66                              & 0.70                                & 0.88                                 & 14.9                \\
H-Softmax (nodes)                    & 0.73                              & 0.85                                & \textbf{0.99}                        & 11.3                & 0.64                              & 0.66                                & 0.86                                 & 17.6                \\ \midrule
Ensemble: ML+S                       & \textbf{0.81}                     & \textbf{0.88}                       & \textbf{0.99}                        & \textbf{3.8}        & 0.67                              & 0.70                                & 0.89                                 & 27.2                \\ \midrule
CLIP zero-shot                       & 0.35                              & 0.37                                & 0.66                                 & 66.7                & 0.20                              & 0.20                                & 0.46                                 & 143.4               \\
CLIP 1 tag             & 0.64                              & 0.71                                & 0.94                                 & 13.2                & \textbf{0.74}                     & \textbf{0.75}                       & 0.90                                 & \textbf{9.6}        \\
CLIP all tags                       & 0.61                              & 0.65                                & 0.88                                 & 26.4                & 0.51                              & 0.57                                & 0.87                                 & 48.2                \\
CLIP 1 tag prompt          & 0.70                              & 0.76                                & 0.93                                 & 14.1                & {\ul 0.69}                        & {\ul 0.72}                          & {\ul 0.91}                           & {\ul 10.1}          \\ \midrule
Ens: ML+S+CLIP                  & {\ul 0.80}                        & {\ul 0.87}                          & \textbf{0.99}                        & 5.0                 & 0.68                              & 0.70                                & \textbf{0.92}                        & 13.1 \\
\bottomrule
\end{tabular}
\end{adjustbox}
\caption{Summary of results provided in Tables \ref{tab:test_id} and \ref{tab:test_ood}. We show the mean value of each evaluation metric calculated across all facets. R-Prec stands for R-Precision, AvgRP for Average Rank Position.}
\label{tab:summarized}
\end{table}

Overall, while our models are far from achieving perfect annotation capabilities -- especially on challenging edge cases-- they provide significant value in assisting expert catalogers. By guiding them towards broader categories within the class hierarchy, these models enable experts to concentrate on manual annotation of the finer details, thereby alleviating the need for exhaustive scrutiny of the entire hierarchy. We further explore this idea with the design of the annotation assistant tool presented in section~\ref{sec:tool}.

\subsection{Qualitative results}
\label{sec:qualitative}
\begin{figure}[]
\centering
\begin{tabular}{p{0.45\textwidth}cp{0.45\textwidth}}
\toprule
\includegraphics[width=0.45\textwidth]{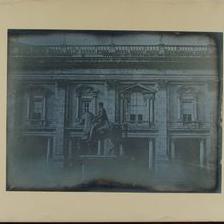} && 
  \includegraphics[width=0.45\textwidth]{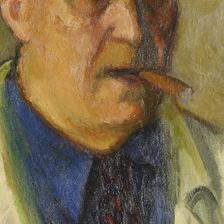} \\ 
  \hline
\textbf{ObjectTypes (Ground-truth):} architectural element, structural element, column, architectural element, opening, window, construction division, room and space, balcony, landscape, architecture
  && \textbf{ObjectTypes (Ground-truth):}
  art, portrait, self-portrait, painting, easel painting \\
\textbf{ObjectTypes(Top-K predictions):} 
  \textcolor{teal}{architectural element}, \textcolor{teal}{structural element}, \textcolor{teal}{architecture}, \textcolor{teal}{opening}, \textcolor{teal}{window, structure}, \textcolor{teal}{column}, \textcolor{orange}{structural element of closed}, \textcolor{orange}{facade}, \textcolor{orange}{building}, ... &&
\textbf{ObjectTypes(Top-K predictions):}  \textcolor{teal}{painting}, \textcolor{teal}{art}, \textcolor{teal}{clothing}, \textcolor{teal}{easel painting}, \textcolor{red}{main garment}, \textcolor{red}{traditional dress}, \textcolor{red}{garment}, \textcolor{orange}{dress accessory}, \textcolor{orange}{figure}, \textcolor{teal}{portrait}, ...\\ 
  \midrule
\textbf{Materials (Ground-truth):}
  metal, non-ferrous metal, copper alloy, bronze,  tile && \textbf{Materials (Ground-truth):}
  paper, cardboard, cardstock \\
\textbf{Materials (Top-K predictions):}   \textcolor{teal}{tile}, \textcolor{red}{water}, \textcolor{orange}{rock}, \textcolor{orange}{metamorphic rock}, \textcolor{orange}{marble}, \textcolor{teal}{non-ferrous metal}, \textcolor{teal}{metal}, \textcolor{teal}{bronze}, \textcolor{teal}{copper alloy}, \textcolor{orange}{glass}, ... && \textbf{Materials (Top-K predictions):}
  \textcolor{teal}{paper}, \textcolor{teal}{cardboard}, \textcolor{teal}{cardstock}, ... \\ 
  \midrule
\textbf{Disciplines (Ground-truth):} 
  architecture && \textbf{Disciplines (Ground-truth):}
  fine arts \\
\textbf{Disciplines (Top-K predictions):} 
  \textcolor{teal}{architecture}, ... &&
 \textbf{Disciplines (Top-K predictions):}  \textcolor{teal}{fine arts}, ... \\ 
  \midrule
\textbf{Subjects (Ground-truth):} 
  landscape &&
\textbf{Subjects (Ground-truth):}   self-portrait \\
\textbf{Subjects (Top-K predictions):} 
  \textcolor{teal}{landscape}, ... && \textbf{Subjects (Top-K predictions):} 
  \textcolor{orange}{portrait}, \textcolor{teal}{self-portrait}, ... \\
  \bottomrule
\end{tabular}

\caption{Qualitative outcomes of the top-performing baseline on two sample images. Top-K model predictions are colored to indicate \textcolor{teal}{correct}, \textcolor{red}{incorrect}, and \textcolor{orange}{plausible} tags.}
\label{tab:qualitatives}
\end{figure}

In Figure \ref{tab:qualitatives}, we present the qualitative outcomes for two selected images obtained using our top-performing model, which is an ensemble of the methods (ML+S+CLIP) explained earlier. Correct predictions are highlighted in green, predictions that, while not annotated in the ground truth but reasonable, are highlighted in orange, and irrelevant tangs are highlighted in red. It is worth noting that the model exhibits consistency in its predictions, although ambiguities occur more often than not at the annotation level. For instance, in the right image, the prediction of the  ``portrait'' tag for the ``Subject'' facet is reasonable, but not part of the ground-truth annotations of this image. This illustrates the difficulty in evaluating image tagging models for real-world applications.

\subsection{An annotation assistant tool for GLAM cataloguers}
\label{sec:tool}

To further explore the practical application of our models in real-world settings, we have developed an annotation assistant tool specifically designed for GLAM  cataloguers. This tool is engineered to leverage the strengths of the ensemble model (ML+S+CLIP), providing an intuitive and interactive interface for image annotation tasks.

The core functionality of the tool is to present the hierarchical tag structure for each facet involved in the cataloguing process. Upon uploading an image, the tool dynamically updates the tree hierarchies based on the predictions from the ensemble model. Notably, it highlights the nodes within the top 10 predictions for each facet, guiding cataloguers to the most likely categories at various granularity levels. This feature is crucial, as it allows users to visually explore different branches of the class hierarchy efficiently, thereby facilitating a focused review of the relevant categories.

\begin{figure}
    \centering
    \includegraphics[width=\textwidth]{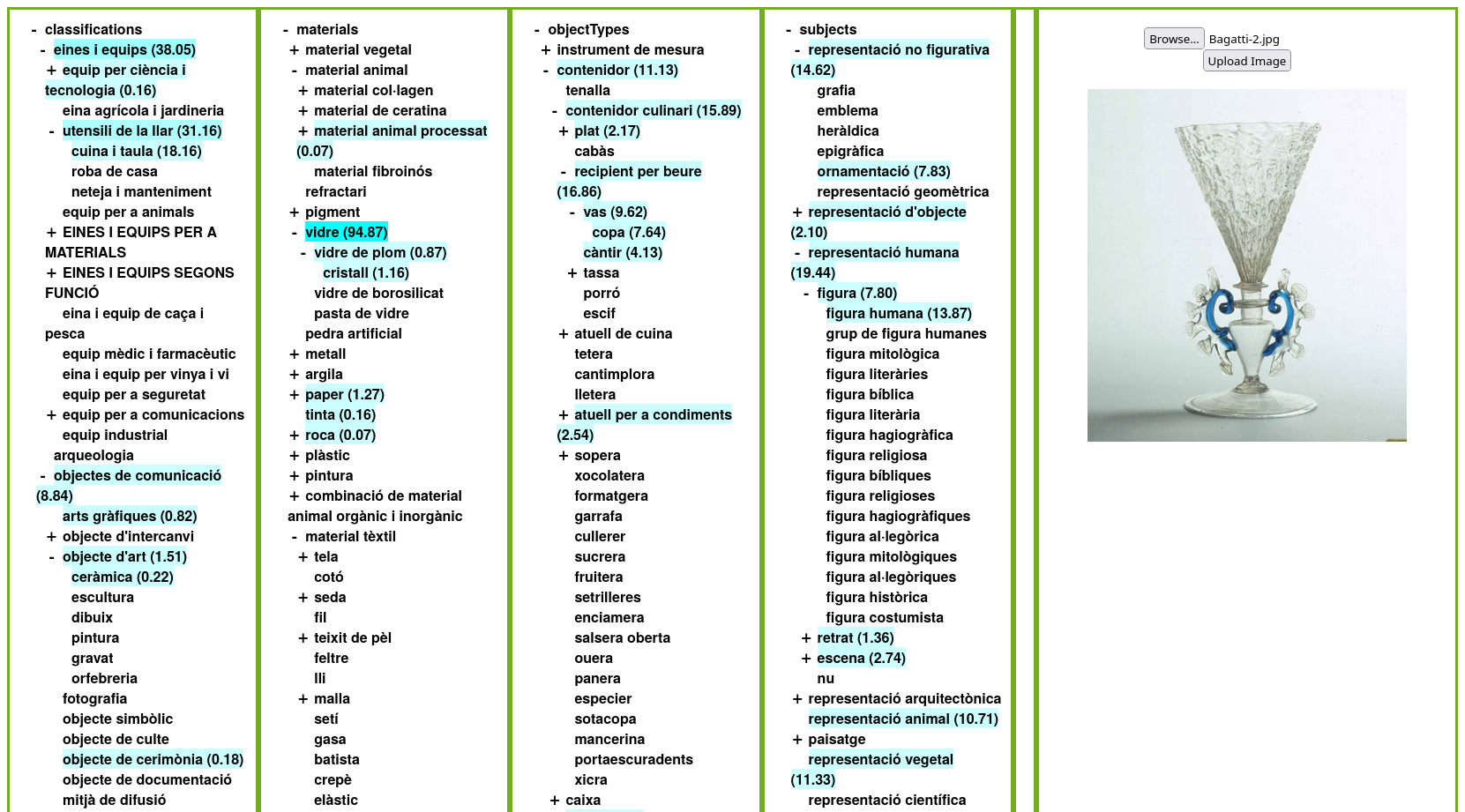}
    \caption{Annotation Assistant Tool Interface. This screenshot displays the tool's functionality where the hierarchical class structure for each facet is shown alongside an uploaded image. The top 10 predictions are highlighted (based on their relevance score) within the hierarchies, enabling GLAM cataloguers to explore various classification branches and focus on detailed annotations. Notice that branches without a highlighted relevant child can still be explored by expanding them with a mouse click on the + symbol. Note: tag labels are shown in Catalan because it is the language of the cataloguers that have been testing the tool. }
    \label{fig:tool_screenshot}
\end{figure}

Figure \ref{fig:tool_screenshot} shows a screenshot of the tool in action. The image depicts how the tree hierarchies are displayed alongside the uploaded image, with the top predictions highlighted. This visualization not only aids in quick navigation through the class hierarchy but also helps in understanding the model's reasoning process, making it a valuable resource for cataloguers. 

By streamlining the initial stages of annotation, the tool enables experts to concentrate on refining the annotations at finer detail levels. This approach significantly reduces the time and effort required for exhaustive manual scrutiny of every possible category, allowing for more efficient allocation of human expertise where it is most needed. The utility of the tool has been validated by domain experts.

\section{Conclusion}
\label{sec:conclusion}
Our study into automatic metadata annotation for GLAM institutions reveals the significant potential of the EUFCC-340K dataset to enhance cataloging efficiency. By leveraging a hierarchical annotation system, our approach addresses the long-tail distribution challenge and incomplete annotations common in cultural heritage assets. Experimental results confirm the efficacy of our baseline models, particularly when using an ensemble method. Although the performance on the Outer test set indicates challenges in generalization, our study lays a foundational framework for future research aimed at further improving annotation aid tools in the GLAM domain. In hopes of fostering further research within the community, we have made the EUFCC-340K dataset publicly available at \url{https://github.com/cesc47/EUFCC-340K}.


\section*{Declarations}

\begin{itemize}
    \item \textbf{Funding} \\
    This work has been supported by the ACCIO INNOTEC 2021 project Coeli-IA (ACE034/21/000084), and the CERCA Programme / Generalitat de Catalunya. Lluis Gomez is funded by the Ramon y Cajal research fellowship RYC2020-030777-I / AEI / 10.13039/501100011033.

    \item \textbf{Conflict of interest/Competing interests} \\
    The authors declare the following financial interests/personal relationships which may be considered as potential competing interests: This work was supported by the ACCIO INNOTEC 2021 project Coeli-IA (ACE034/21/000084) and the CERCA Programme/Generalitat de Catalunya. Lluis Gomez is funded by the Ramon y Cajal research fellowship RYC2020-030777-I / AEI / 10.13039/501100011033.

    \item \textbf{Ethics approval and consent to participate} \\
    Not applicable

    \item \textbf{Consent for publication} \\
    Not applicable

    \item \textbf{Data availability} \\
    The dataset generated during and analyzed during the current study is available in the EUFCC-340K repository, accessible via \url{https://github.com/cesc47/EUFCC-340K}. This dataset is made publicly available to foster further research within the community. The dataset has been collected using the Europeana REST API\footnote{\url{https://pro.europeana.eu/page/intro}} and includes only records tagged with ``REUSABILITY:open,'' ensuring that the dataset comprises images suitable for open research and application\footnote{\url{https://pro.europeana.eu/post/can-i-use-it}}. 

    \item \textbf{Materials availability} \\
    Not applicable

    \item \textbf{Code availability} \\
    Not applicable

    \item \textbf{Author contribution} \\
All authors contributed to the material preparation, study conception, and design. Data collection and analysis were performed by Francesc Net, Marc Folia, and Lluis Gomez. Francesc Net and Lluis Gomez were responsible for the baseline models' design and implementation. All authors actively participated in the analysis and discussion of the results, ensuring a comprehensive evaluation of the study findings. The first draft of the manuscript was written by Francesc Net, and all authors contributed to drafting and revising the manuscript. All authors have read and approved the final manuscript.

\end{itemize}

\bibliography{sn-bibliography}

\end{document}

%% file: tikz_samples2.tex
\begin{figure}
  \centering
  \setlength{\tabcolsep}{1pt}
  \begin{tabular}{cc}
  \toprule 

  \includegraphics[height=0.29\linewidth]{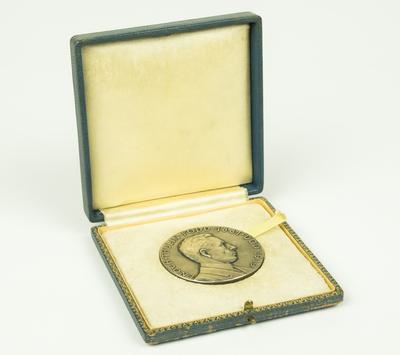} &
\begin{tikzpicture}[node distance=1em]

\node (ObjectType) [objects, xshift=4em] {\tiny{ObjectType}};
\node (type1) [objects, below of=ObjectType, yshift=-0.85em, xshift=-2em] {\tiny{Medal}};
\node (type2) [objects, below of=ObjectType, yshift=-0.85em, xshift=2em] {\tiny{Textile}};

\node (Material) [materials, right of= ObjectType, xshift=12em] {\tiny{Material}};
\node (mat1) [materials, below of= Material, yshift=-0.85em, xshift=-5.5em] {\tiny{Animal material}};
\node (mat2) [materials, below of= mat1, yshift=-1em,text width=20mm] {\tiny{\baselineskip=2pt processed animal material\par}};
\node (mat3) [materials, below of= mat2, yshift=-1em] {\tiny{Leather}};
\node (mat4) [materials, below of= mat3, yshift=-0.85em] {\tiny{Suede}};

\node (mat5) [materials, below of= Material, yshift=-0.85em, xshift=0em] {\tiny{Paper}};
\node (mat6) [materials, below of= mat5, yshift=-0.85em] {\tiny{Cardboard}};
\node (mat7) [materials, below of= mat6, yshift=-0.85em] {\tiny{Cardstock}};

\node (mat8) [materials, below of= Material, yshift=-0.85em, xshift=5em] {\tiny{Metal}};
\node (mat9) [materials, below of= mat8, yshift=-1em,text width=15mm] {\tiny{\baselineskip=2pt Non-ferreous metal\par}};
\node (mat10) [materials, below of= mat9, yshift=-1em] {\tiny{Silver}};

\draw [arrow] (ObjectType) -- (type1);
\draw [arrow] (ObjectType) -- (type2);

\draw [arrow] (Material) -- (mat1);
\draw [arrow] (mat1) -- (mat2);
\draw [arrow] (mat2) -- (mat3);
\draw [arrow] (mat3) -- (mat4);

\draw [arrow] (Material) -- (mat5);
\draw [arrow] (mat5) -- (mat6);
\draw [arrow] (mat6) -- (mat7);

\draw [arrow] (Material) -- (mat8);
\draw [arrow] (mat8) -- (mat9);
\draw [arrow] (mat9) -- (mat10);
\end{tikzpicture}
\\

\midrule

\includegraphics[width=0.35\linewidth]{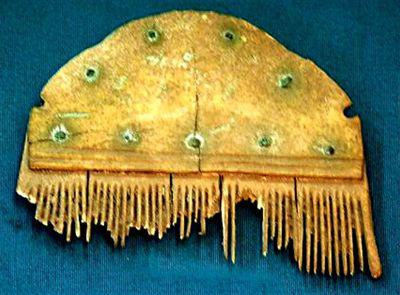} &
\begin{tikzpicture}[node distance=1em]

\node (ObjectType) [objects, xshift=4em] {\tiny{ObjectType}};
\node (type1) [objects, below of=ObjectType, yshift=-0.85em, xshift=-2em] {\tiny{Pottery}};
\node (type2) [objects, below of=ObjectType, yshift=-0.85em, xshift=2em] {\tiny{Clothing}};
\node (type3) [objects, below of=type2, yshift=-0.85em] {\tiny{Dress accessory}};
\node (type4) [objects, below of=type3, yshift=-0.85em] {\tiny{Hair accessory}};
\node (type5) [objects, below of=type4, yshift=-0.85em] {\tiny{Hair ornament}};

\node (Material) [materials, right of= ObjectType, xshift=8em] {\tiny{Material}};
\node (mat1) [materials, below of= Material, yshift=-0.85em] {\tiny{Clay}};
\node (mat2) [materials, below of= mat1, yshift=-0.85em] {\tiny{Ceramics}};

\node (Discipline) [discipline, right of= Material, xshift=6em] {\tiny{Discipline}};
\node (dis1) [discipline, below of= Discipline, yshift=-0.85em] {\tiny{Archaeology}};

\draw [arrow] (ObjectType) -- (type1);
\draw [arrow] (ObjectType) -- (type2);

\draw [arrow] (Material) -- (mat1);
\draw [arrow] (mat1) -- (mat2);
\draw [arrow] (type2) -- (type3);
\draw [arrow] (type3) -- (type4);
\draw [arrow] (type4) -- (type5);
\draw [arrow] (Discipline) -- (dis1);
\end{tikzpicture}
\\

\midrule 

\includegraphics[width=0.35\linewidth]{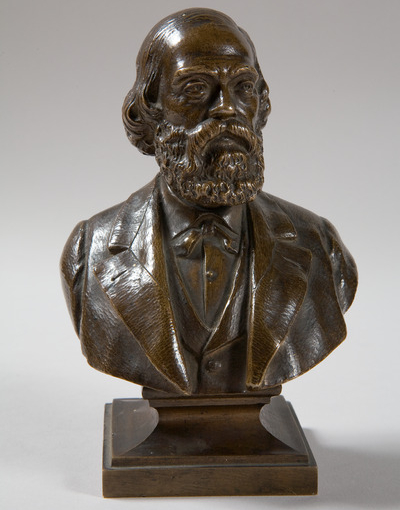} &
\begin{tikzpicture}[node distance=1em]

\node (ObjectType) [objects, xshift=4em] {\tiny{ObjectType}};
\node (type1) [objects, below of=ObjectType, yshift=-0.85em, xshift=-2em] {\tiny{Figure}};
\node (type2) [objects, below of=ObjectType, yshift=-0.85em, xshift=2em] {\tiny{Sculpture}};
\node (type3) [objects, below of=type1, yshift=-0.85em] {\tiny{Bust}};

\node (Material) [materials, right of= ObjectType, xshift=7em] {\tiny{Material}};
\node (mat8) [materials, below of= Material, yshift=-0.85em] {\tiny{Metal}};
\node (mat9) [materials, below of= mat8, yshift=-1em,text width=15mm] {\tiny{\baselineskip=2pt Non-ferreous metal\par}};
\node (mat10) [materials, below of= mat9, yshift=-1em] {\tiny{Coper alloy}};
\node (mat11) [materials, below of= mat10, yshift=-1em] {\tiny{Bronze}};

\node (empty) [below of= mat11, yshift=-5em] {};

\node (Subject) [subjects, right of= Material, xshift=6em] {\tiny{Subject}};
\node (sub1) [subjects, below of= Subject, yshift=-0.85em] {\tiny{Figure}};

\draw [arrow] (ObjectType) -- (type1);
\draw [arrow] (ObjectType) -- (type2);
\draw [arrow] (type1) -- (type3);

\draw [arrow] (Material) -- (mat8);
\draw [arrow] (mat8) -- (mat9);
\draw [arrow] (mat9) -- (mat10);
\draw [arrow] (mat10) -- (mat11);

\draw [arrow] (Subject) -- (sub1);
\end{tikzpicture}
\\

\midrule 

\includegraphics[width=0.35\linewidth]{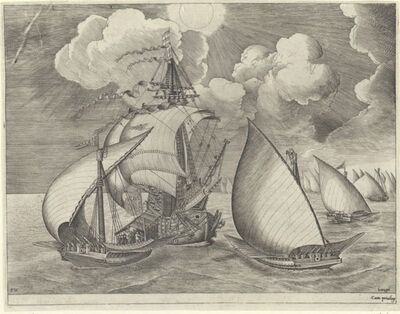} &
\begin{tikzpicture}[node distance=1em]

\node (ObjectType) [objects, xshift=4em] {\tiny{ObjectType}};
\node (type1) [objects, below of=ObjectType, yshift=-0.85em, xshift=-5em] {\tiny{Transport vehicle}};
\node (type2) [objects, below of=ObjectType, yshift=-0.85em, xshift=5em] {\tiny{Stamp}};

\node (type3) [objects, below of=ObjectType, yshift=-0.85em] {\tiny{Seascape}};

\node (type4) [objects, below of=type1, yshift=-0.85em] {\tiny{Vehicle}};
\node (type5) [objects, below of=type4, yshift=-0.85em] {\tiny{Boat}};
\node (type6) [objects, below of=type5, yshift=-0.85em] {\tiny{Cruise}};
\node (type7) [objects, below of=type6, yshift=-0.85em] {\tiny{Warship}};

\node (type8) [objects, below of=type2, yshift=-0.85em] {\tiny{Stamp}};
\node (type9) [objects, below of=type8, yshift=-0.85em] {\tiny{Intaglio}};
\node (type10) [objects, below of=type9, yshift=-0.85em, xshift=-2em] {\tiny{Etching}};
\node (type11) [objects, below of=type9, yshift=-0.85em, xshift=2em] {\tiny{Engraving}};

\node (Subject) [subjects, right of= ObjectType, xshift=11em] {\tiny{Subject}};
\node (sub1) [subjects, below of= Subject, yshift=-0.85em, xshift=-2em] {\tiny{Landscape}};
\node (sub2) [subjects, below of= Subject, yshift=-0.85em, xshift=2em] {\tiny{Marina}};

\node (Material) [materials, right of= ObjectType, xshift=11em, yshift=-6em] {\tiny{Material}};
\node (mat1) [materials, below of= Material, yshift=-0.85em] {\tiny{Paper}};

\draw [arrow] (Material) -- (mat1);

\draw [arrow] (Subject) -- (sub1);
\draw [arrow] (Subject) -- (sub2);

\draw [arrow] (ObjectType) -- (type1);
\draw [arrow] (ObjectType) -- (type2);
\draw [arrow] (ObjectType) -- (type3);

\draw [arrow] (type1) -- (type4);
\draw [arrow] (type4) -- (type5);
\draw [arrow] (type5) -- (type6);
\draw [arrow] (type6) -- (type7);

\draw [arrow] (type2) -- (type8);
\draw [arrow] (type8) -- (type9);
\draw [arrow] (type9) -- (type10);
\draw [arrow] (type9) -- (type11);
\end{tikzpicture}
\\

\bottomrule
  \end{tabular}
     
  \caption{Sample records from the EUFCC-340K dataset. Each image in the dataset is annotated across different facets' hierarchies of the Getty ``Art \& Architecture Thesaurus''. Some nodes were omitted for visualization purposes.}
  \label{image:europeana_labeled}
\end{figure}